\documentclass[sigconf]{acmart}

\def\BibTeX{{\rm B\kern-.05em{\sc i\kern-.025em b}\kern-.08emT\kern-.1667em\lower.7ex\hbox{E}\kern-.125emX}}

\usepackage{subfigure}
\usepackage{multirow}
\usepackage{graphicx}
\begin{document}

\copyrightyear{2019}
\acmYear{2019}
\setcopyright{usgovmixed}
\acmConference[GLSVLSI '19]{Great Lakes Symposium on VLSI 2019}{May 9--11, 2019}{Tysons Corner, VA, USA}
\acmBooktitle{Great Lakes Symposium on VLSI 2019 (GLSVLSI '19), May 9--11, 2019, Tysons Corner, VA, USA}
\acmPrice{15.00}
\acmDOI{10.1145/3299874.3319493}
\acmISBN{978-1-4503-6252-8/19/05}

%
\title{On the use of Deep Autoencoders for Efficient Embedded Reinforcement Learning}

%
%

\author{Bharat Prakash}
\affiliation{%
  \institution{University of Maryland \\
  Baltimore County}}

\author{Mark Horton}
\affiliation{
  \institution{University of Maryland\\
  Baltimore County}}

\author{Nicholas R. Waytowich}
\affiliation{
  \institution{US Army Research Laboratory}}

\author{William David Hairston}
\affiliation{
  \institution{US Army Research Laboratory}}

\author{Tim Oates}
\affiliation{
  \institution{University of Maryland \\
  Baltimore County}}

\author{Tinoosh Mohsenin}
\affiliation{
  \institution{University of Maryland\\
  Baltimore County}}

%
\renewcommand{\shortauthors}{}

%
\begin{abstract}
In autonomous embedded systems, it is often vital to reduce the amount of actions taken
in the real world and energy required to learn a policy. Training reinforcement learning agents from high dimensional image representations can be very expensive and time consuming. Autoencoders are deep neural network used to compress high dimensional data such as pixelated images into small latent representations. This compression model is vital to efficiently learn policies, especially when learning on embedded systems. We have implemented this model on the NVIDIA Jetson TX2 embedded GPU, and evaluated the power consumption, throughput, and
energy consumption of the autoencoders for
various CPU/GPU core combinations, frequencies, and model parameters. Additionally, we
have shown the reconstructions generated by the autoencoder to analyze the quality of the generated
compressed representation and also the performance of the reinforcement learning agent. Finally, we have presented an assessment of the viability of training these models on embedded systems and their usefulness in developing autonomous policies. Using autoencoders, we were able to achieve 4-5 $\times$ improved performance compared to a baseline RL agent with a convolutional feature extractor, while using less than 2W of power.

\end{abstract}

%
%

%


\maketitle

\section{Introduction}

Recently, deep reinforcement learning algorithms have provided an effective approach to automating tasks which previously could only be performed by humans.  The most popular modern reinforcement learning algorithms use deep neural networks and back-propagation in order to maximize a reward function which indicates how well the agent is performing in the environment.  However, when these algorithms are applied to high-dimensional data such as pixelated images, they implicitly must learn to extract useful features from an environment.  This feature extraction requires the training of large convolutional layers, which learn to detect shapes in the training samples.  Deep convolutional networks used in tasks such as reinforcement learning and image classification are notoriously power intensive, requiring optimization and specialized hardware in order to be viable for embedded applications \cite{jafari2018sensornet}.  Training these networks is time intensive and often requires very large datasets.  When training reinforcement learning algorithms for real life tasks such as autonomous driving, it is impractical to collect large datasets of images while taking the actions suggested by a partially trained reinforcement learning algorithm, which is likely to lead to crashes and costly outcomes.  However, learning to detect important visual features does not necessarily require taking undirected actions, and can happen outside of the context of reinforcement learning.  For this reason, an autoencoder can be used to compress high-dimensional visual inputs into a more compressed representation which can be used in a smaller network that requires far fewer iterations to learn an  effective policy. 

In order to preserve the safety of our agent and expedite the training of the autoencoder, we are using a methodology in which a large number of samples are gathered by a human expert, exported to a powerful GPU server, and used to train an autoencoder which is then imported onto the embedded GPU and used to perform compression and reinforcement learning in real time.  The authors of "Learning to Drive Smoothly in Minutes" and the associated github repository \cite{drive-smoothly-in-minutes} provided an implementation of this methodology in the Donkey Car environment, which we will use to demonstrate its potential viability in embedded real-world tasks. For example, if we were training a self-driving car, we would have a human driving and collecting video, export the video to a GPU server, train an autoencoder to compress and reconstruct the video frames, import the encoder parameters back into the self-driving car's embedded GPU, and then we would perform standard reinforcement learning on the compressed version of images captured by the camera.  This way, the autonomous car would learn a driving policy significantly faster and with significantly fewer crashes than if we were to train a deep convolutional neural network end to end.  In this paper, we will 
demonstrate the viability of using this architecture in real time on an embedded system to choose an optimal action from frames of video captured by the camera.  Additionally, we will vary different parameters in our encoder in an attempt to reduce its computational cost, increase the accumulated reward of the policy network, and decrease the number of iterations needed to achieve a satisfactory reward.  We believe that this architecture presents an effective approach to performing faster and safer reinforcement learning on embedded devices.  We also believe that by optimizing our autoencoder architecture, we can reduce the amount of power and time needed to reach a successful policy, and we can improve the peak performance of our agent.

\section{Related Work}

Over the past several years, there has been significant interest in performing reinforcement learning in high-dimensional visual environments. This gained traction starting with Deepmind's landmark paper in which they were able to successfully play Atari games using pixelated images as the input to a Deep Q-Network \cite{mnih2015human}.  Since then, there has been promise that self-driving cars, autonomous drones, and smart robotic systems would be soon to follow.  However, despite innovations to the underlying reinforcement learning algorithms \cite{sutton2000policy}, \cite{mnih2016asynchronous}, \cite{schulman2017proximal}, engineers have often struggled to apply these algorithms to real-life domains.  

\begin{figure}
		\centering{\includegraphics[width=0.45\textwidth]{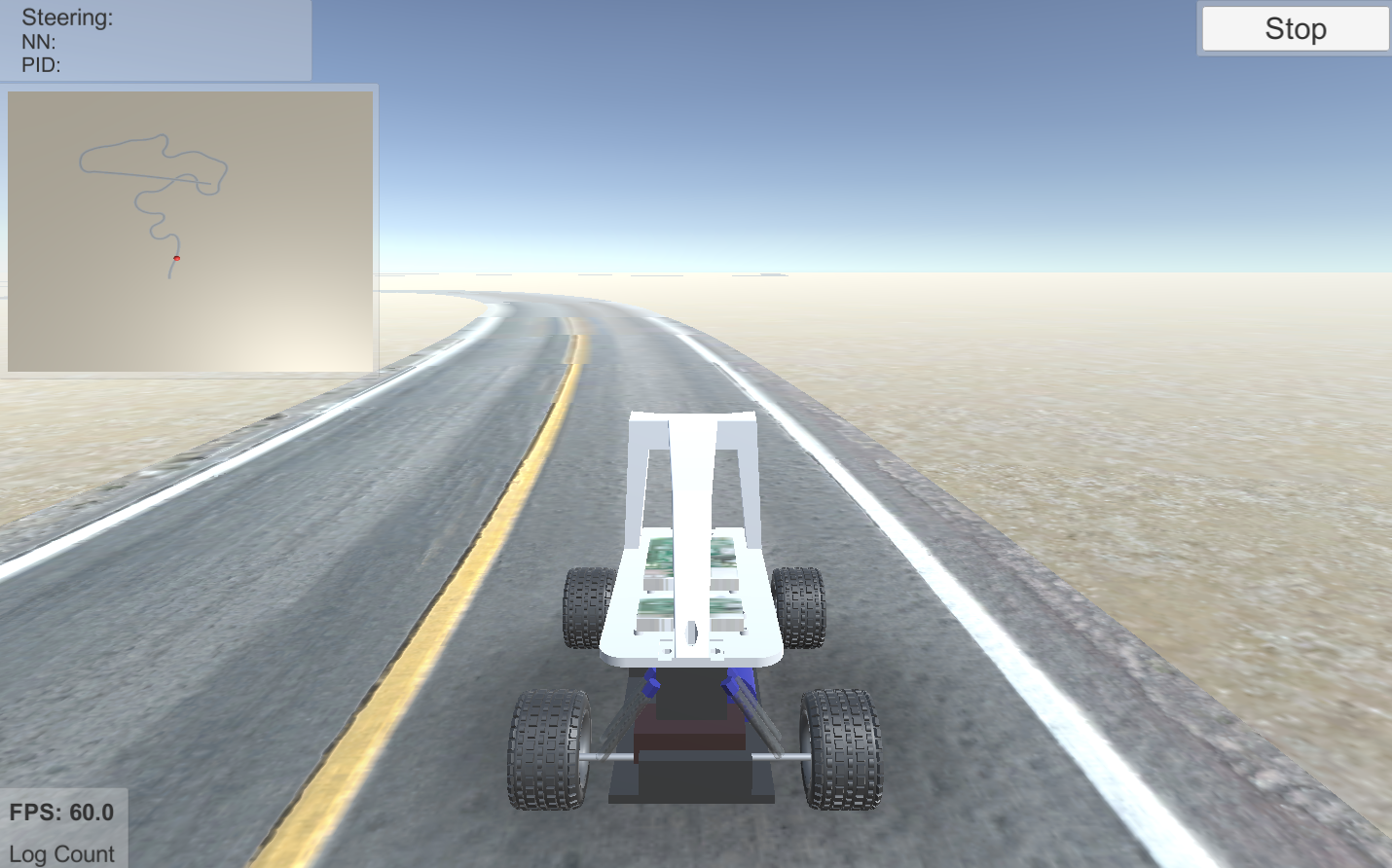}}
		\caption{Donkey Car Environment}
		\label{donkey_env} 
\end{figure}

Modern reinforcement learning algorithms require vast amounts of data to learn policies in complex environments, resulting in costly real-world actions. The amount of data needed varies between different reinforcement learning algorithms, so we are using the Soft Actor Critic (SAC) \cite{haarnoja2018soft} algorithm used in \cite{drive-smoothly-in-minutes} for its ability to learn high-quality policies with a small number of samples. This will be of vital importance when training on a sensitive embedded system.  Unlike many other state of the art reinforcement learning algorithms, SAC is off-policy, meaning that it can learn from a random set of environment interactions, instead of a specific sequential set as is required by on-policy algorithms.  This is critical in embedded environments, because on-policy algorithms are bottlenecked by the rate at which an agent can take actions, which is often much lower in physical systems such as autonomous vehicles than in simulated environments such as videogames. Additionally, the ability to learn from previously taken actions improves the sample efficiency of this algorithm, meaning that fewer dangerous steps need to be taken by the agent in order to learn an effective policy.  Finally, this algorithm has a high degree of stability, meaning that there is a low variation in the reward achieved during different training cycles, making it ideal for performance tests.

\begin{figure}
		\centering{\includegraphics[width=0.45\textwidth]{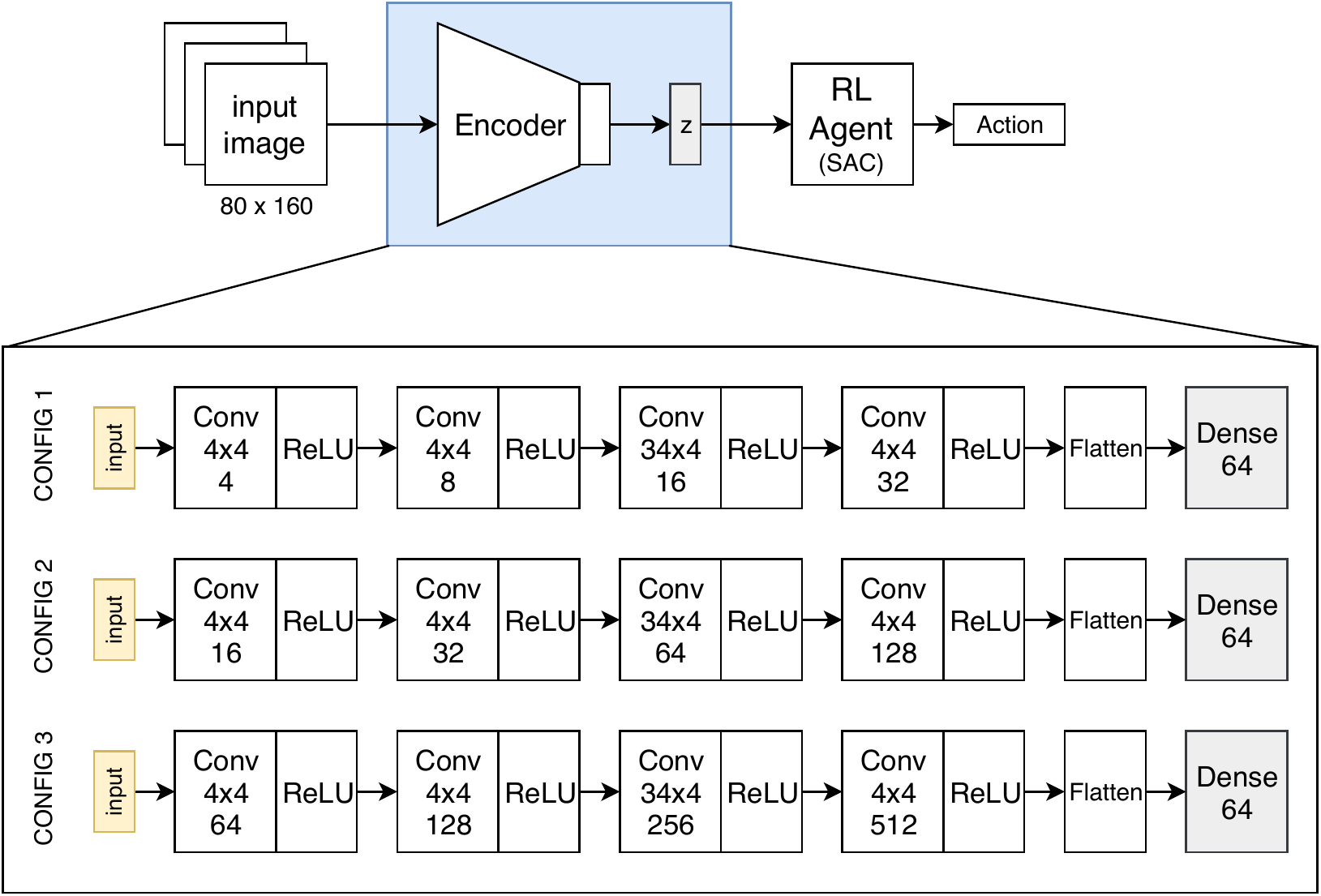}}
		\caption{Proposed Architecture - The input image is encoded using the pre-trained encoder to obtain the latent representation. This is then used by the Soft Actor Critic RL agent to output the actions. It also shows the 3 different configurations of the encoder with increasing trainable parameters}
		\label{fig:arch} 
\end{figure}

Training the deep convolutional networks commonly used for reinforcement learning is often power and time intensive, requiring substantial hardware and offline training.  In order to address these issues, engineers have often opted to produce high-fidelity simulations in order to train networks offline , or explicitly extract a small set of meaningful, hand-crafted features from a high dimensional environment so that training can be done quickly on a smaller network online.  These approaches can expand reinforcement learning to more domains, but producing simulations and manually extracting features both require human labor and expertise, and often fail to generalize to environments with more complex dynamics and features. In these cases when it becomes excessively burdensome to explicitly model environments, autoencoders can replace human experts and learn compressed representations of high dimensional environments without supervision \cite{ha2018world}.  These compressed representations can either be used to produce simulations and train offline  \cite{ha2018world}, or train smaller networks online learning from compressed representations \cite{Warnell2018, Waytowich2018, Goecks2018}. While it is not a novel idea to use autoencoders to accelerate reinforcement learning, we have not found substantial work focused applying these models to embedded learning tasks where power and environment interactions are valuable, limited resources.

An autoencoder is a simple concept: a neural network which is trained to output a close approximation of its input. Generally, these models consist of an encoder, which maps a high dimensional input such as an image into a reduced vector of latent variables, and a decoder, which uses this reduced representation to generate a reconstruction of the original high-dimensional input.  There are many variations of this model, but we will focus on the Variational Autoencoders (VAE) in particular. Variational Autoencoders have been used in a number of reinforcement learning tasks for its ability to do substantial unsupervised compression and partial disentanglement \cite{ha2018world}. The variational autoencoder is an autoencoder in which guassian noise is added to the latent space, and each variable in the latent space is incentivized to conform to a multidimensional gaussian distribution with predetermined parameters.

Although they can create small, easy to work with representations of large input data, convolutional autoencoders themselves are generally large, power-intensive models.  While one can avoid much of this burden by training the autoencoder offline and leaving the only online training for the policy network, it is still necessary to perform inference on the encoder at each time step in order to compress the sample.  Therefore, we will also explore variations we can make to the encoder architecture and the corresponding effects it will have on the performance of the reinforcement learning algorithm, the energy required to encode a frame of video, and the latency associated with this encoding.  
We will also explore the effect of the number of filters on all of these performance parameters. Using this information, we will analyse the viability of this approach to information compression and reinforcement learning on an embedded Nvidia Jetson TX2 platform.

\section{Methods}

In this section, we will explain the architecture of our autoencoder and reinforcement learning algorithm. This includes a description of the environment in Section 3.1., the collecton of training data for the autoencoder in section 3.2, and the training of the RL agent in Section 3.3.

\subsection{Environment}
We have used the Donkey Car Simulator \cite{drive-smoothly-in-minutes} as our test environment in which our RL agent acts. This simulator, shown in Figure \ref{donkey_env}, is built on the the Unity game platform and can be easily interacted with via python libraries. The agent is a toy car with the goal of maximizing speed while staying on the track. The input to the agent is a 120x160 image from the car's camera and the action space is [steering, throttle] where these are continuous values. The reward function we get from the environment is a function of cross track error ($cte$). Cross track error measures the distance between the center of the track and car. The shaped reward we use to train the reinforcement learning agent is $reward = 1 - (abs(cte)/max\_cte)*speed$ where $max\_cte$ is a normalizing constant.

\subsection{Pre-Training the Autoencoder}

The first step in the process is to collect data to train the autoencoder. We collected these images using a human driver who controlled the car for a fixed number of steps until we obtained the required number of images. 

We collected 5000 images using the human driver and then trained a variational autoencoder with the objective of re-constructing the original image fed to the network. The autoencoder was trained for 100 epochs.
Performing the task of reconstruction with a bottleneck in the network forces the encoder to compress the original image to a lower dimensional vector in the latent space. We do not need to use the embedded device for this step, so it can be performed offline on a more powerful server.

We repeated this process with various architectures of the variational autoencoders by changing the filter sizes and observing how the performance varies in terms of encoding, power consumption, and the performance of the reinforcement agent trained in the latent space. We show this in more detail in the following sections.

\subsection{Training the RL Agent}

The RL agent is trained using the Soft actor critic(SAC) algorithm with the following reward function:
\begin{itemize}
  \item reward = -10 - w1 x throttle (when getting off track)
  \item reward = +1 + w2 x throttle (when on track)
  \item w1 and w2 are hyperparameters which can be tuned
\end{itemize}
During training, the input images are compressed by the encoder of the variation autoencoder in order to obtain the latent vector. The RL agent, instead of receiving the high dimensional image, will receive the compressed representation from the encoder. This means that the RL agent can be a simple feed forward neural network, which will train much quicker than its larger convolutional counterpart.

This architecture is especially beneficial for embedded devices since we also see lower power consumption while training compared to the full RL agent with a CNN feature extractor.

The basic architecture and the configurations are shown in Figure \ref{fig:arch}. We can see that the input image is passed to the pre-trained encoder which gives us the compressed representation of the image. This is then passed to the RL agent to learn a policy and output the actions. The actions in our case are throttle and steering values. An episode ends when the car goes off track and the weights of the RL agent are updated at the end of each episode. The RL agent is trained for 150 episodes and we repeat this with different configurations of the autoencoder.

\section{Experiments and Results}

This section will detail the experiments we performed with different configurations of the variational autoencoder as well as the results for each configuration. We also show the power consumption and energy efficiency on Nvidia TX2 for different CPU and GPU frequency combinations. 

\begin{figure*}
		\centering{\includegraphics[width=0.95\textwidth]{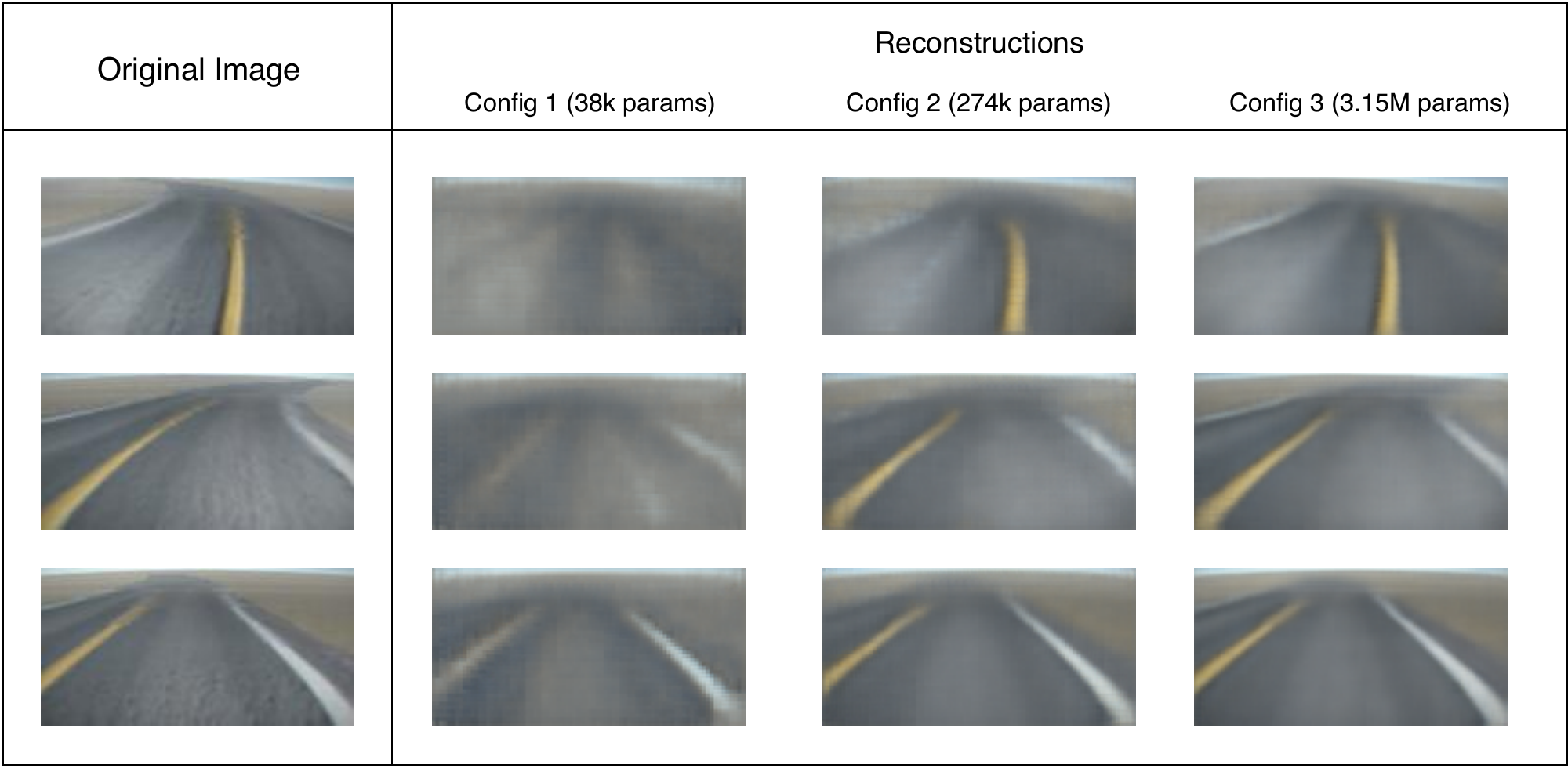}}
		\caption{Autoencoder image reconstructions for different configurations. This shows how reconstruction quality improves as we increase the number of parameters. This implies that we obtain more detailed latent space representations as we increase the number of parameters. }
		\label{fig:recons} 
\end{figure*}


\subsection{VAE configurations}

We experimented with various configurations of the VAE by varying the number of convolutional filters in each layer as shown in Figure \ref{fig:arch}. In the first set of experiments, we changed the number of filters in the convolutional layer thereby increasing the number of trainable parameters from from 38k, 274k to 3.15M. Figure 3 shows how the quality of the reconstruction from the autoencoder changes as we increase the number of parameters. We also measured the quality of reconstruction by calculating Mean Squared Error (MSE) and Peak Signal to Noise Ratio (PSNR) and visually showing reconstructions from each configuration. Table \ref{fig:mse_psnr} shows the MSE and PSNR values for the 3 selected configurations, and Figure 3 shows the reconstructions generated by each VAE.

One can visually observe in Figure 3 that config 3 produced sharper reconstructions than config 2. This is reaffirmed by the the MSE and PSNR metrics in Table \ref{fig:mse_psnr}, which both show quantitatively that the quality of reconstructions increases as the number of parameters in the encoder increases.  Although there is a quantitative difference in reconstruction quality, visual inspection would indicate that config 3 and config 2 both seem to be able to effectively encode the position of the agent on the road as well as the shape of nearby segments of the road.

\subsection{Reinforcement Learning (RL) Performance}


In order to test the effects the various VAE architecures have on policy learning, we trained an RL agent with each of the three pre-trained autoencoders. As a baseline, we also trained the RL agent without a pre-trained autoencoder, using only the raw images, to show the benefits that using an autoencoder has on learning.

\begin{table}[bp]
  \begin{center}
    \begin{tabular}{c|c|c|c} 
    \toprule
      \textbf{}    & \textbf{Config 1} & \textbf{Config 2} & \textbf{Config 3} \\ 
     \textbf{\# Parameters} & \textbf{38 k} & \textbf{274 k*}   & \textbf{3.15 M} \\  
     \midrule
      \textbf{MSE}           &  43.48         & 23.48*           & 18.52  \\ 
      \textbf{PSNR}          &  31.74         & 34.429*          & 36.52  \\
      \bottomrule
    \end{tabular} 
    \caption{Autoencoder Performance for different configurations.}
    \label{fig:mse_psnr}
  \end{center}
\end{table}

This policy network for the baseline RL agent has a CNN feature extractor followed by fully connected layers. In contrast, the policy network in our architecture consists of just fully connected layers and its input is the low dimensional encoded vector from the pre-trained encoder. We measure its performance in terms of episode rewards. The reward is a function of the distance travelled by the car and its speed while staying on the track. 

\begin{figure}[tp]
  \centering{\includegraphics[width=0.45\textwidth]{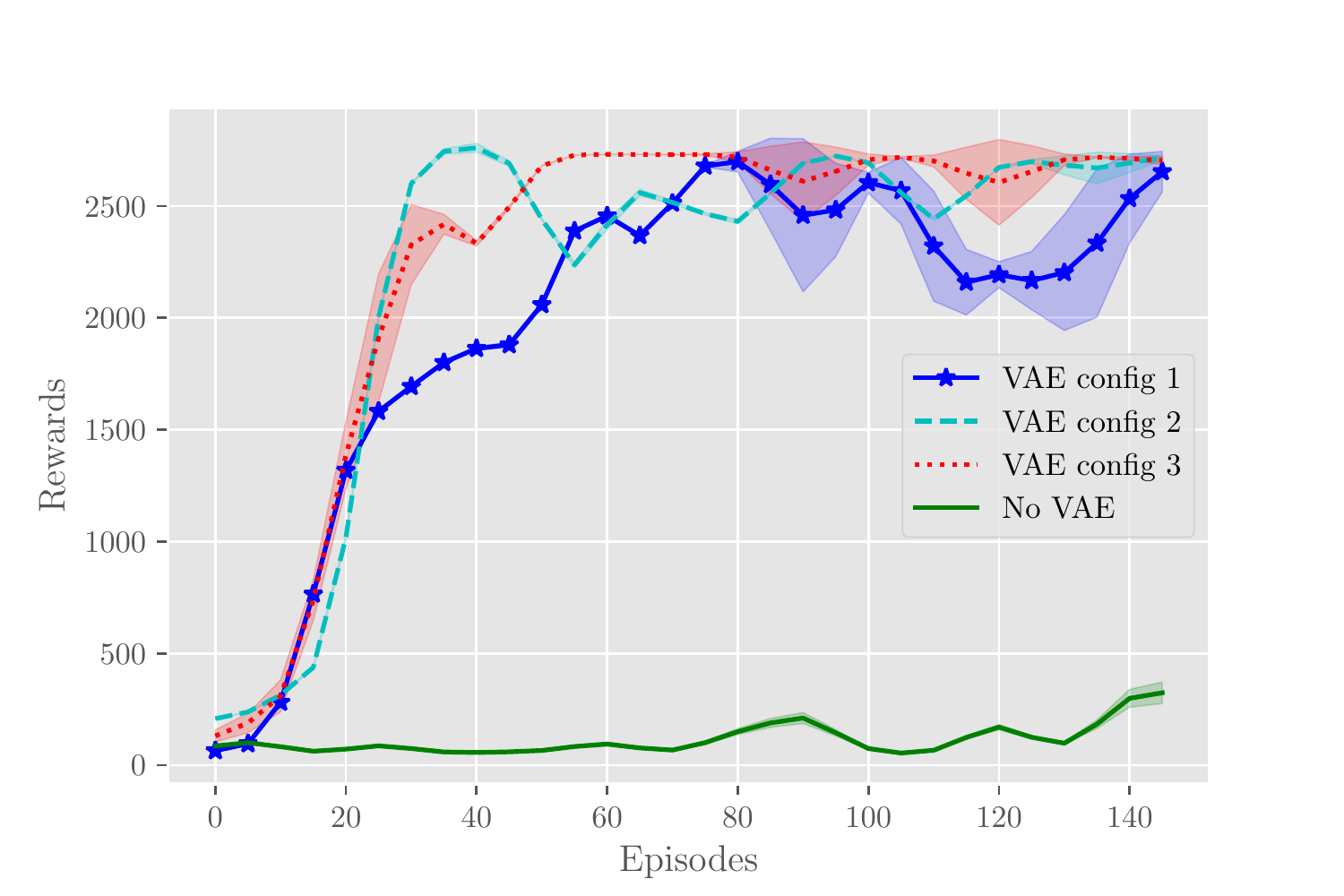}}

\caption{Episode Rewards -  This shows the episode rewards accumulated while training the RL agent using different VAE configurations and also without the VAE. We observe that the RL agent learns considerably faster when trained with pre-trained encoders. Also, we see how the performance varies when we increase the number of convolutional filters in our VAEs.}
    \label{fig:epi_rew}
\end{figure}

We trained the RL agent for 150 episodes and evaluated the agent every 5 episodes, then repeated this process using the three different VAE configurations.
In Figure \ref{fig:epi_rew}, we show the episode rewards for these configurations.  Here, we can observe that the policy learned in the latent space is better and faster than the policy learned in the high dimensional image space (no VAE). When learning from the latent representation, the episode reward reached 2500 after 150 episodes while the baseline RL agent stays around 400. The baseline agent requires more steps to achieve the same performance as our agent. After 50 episodes, the latent space reinforcement learning algorithm was able to produce a viable policy while the standard reinforcement learner made little noticeable progress. There are a few important caveats to this conclusion however.  First of all, the standard reinforcement learning algorithm will generally match or exceed the reward generated by the autoencoder method given a long enough time horizon. Additionally, before the autoencoder can be trained, we must use a human driver to in order to generate a set of training data for the autoencoder.  Finally, the set of information extracted by an autoencoder is necessarily larger than the information needed to perform reinforcement learning, meaning that a raw reinforcement learner may require a smaller architecture with fewer parameters.  However, when using comparable architectures, the reinforcement learner which incorporates an autoencoder can train a viable policy with many fewer dangerous interactions in the environment than with a standard reinforcement learner.

In Figure \ref{fig:epi_rew}, we also demonstrate how changing size of the autoencoder effects the performance of the agent. Over the set of experiments we performed, we found that we could not meaningfully distinguish the performance of config 2 and config 3 in training a reinforcement learning algorithm. The performance of each of these varied between experiments because of stochasticity introduced by retraining the autoencoders and reinforcement learning algorithm from scratch.  However, we found that config 1 consistently took longer to approach it peak reward.  In this case, config 2 and 3 both surpassed a reward of 2000 after roughly 20-30 episodes while config 1 required 40-50 episodes.  
This result seems to be consistent with the detail generated in the reconstructions of each config as show in Figure 3.  In this figure we see slight visual differences in the reconstructions generated by config 2 and config 3.  However, they were both able to effectively encode the position of the agent on the road and the shape of the road, which are the two most important factors to consider when deciding on an action.  However, neither of these are nearly as clear in the config 1 reconstruction, which is likely why the reinforcement learner learning from config 1 embeddings struggled relative to the other two.  Interestingly, even the config 1 reinforcment learner was able to achieve high rewards over a longer time horizon, meaning that even this configuration was encoding substantial information relevant to the agent.

\subsection{Embedded Hardware Metrics}

\begin{figure*}[]

  \centering
  \subfigure[Frame Rate]{\includegraphics[scale=0.33]{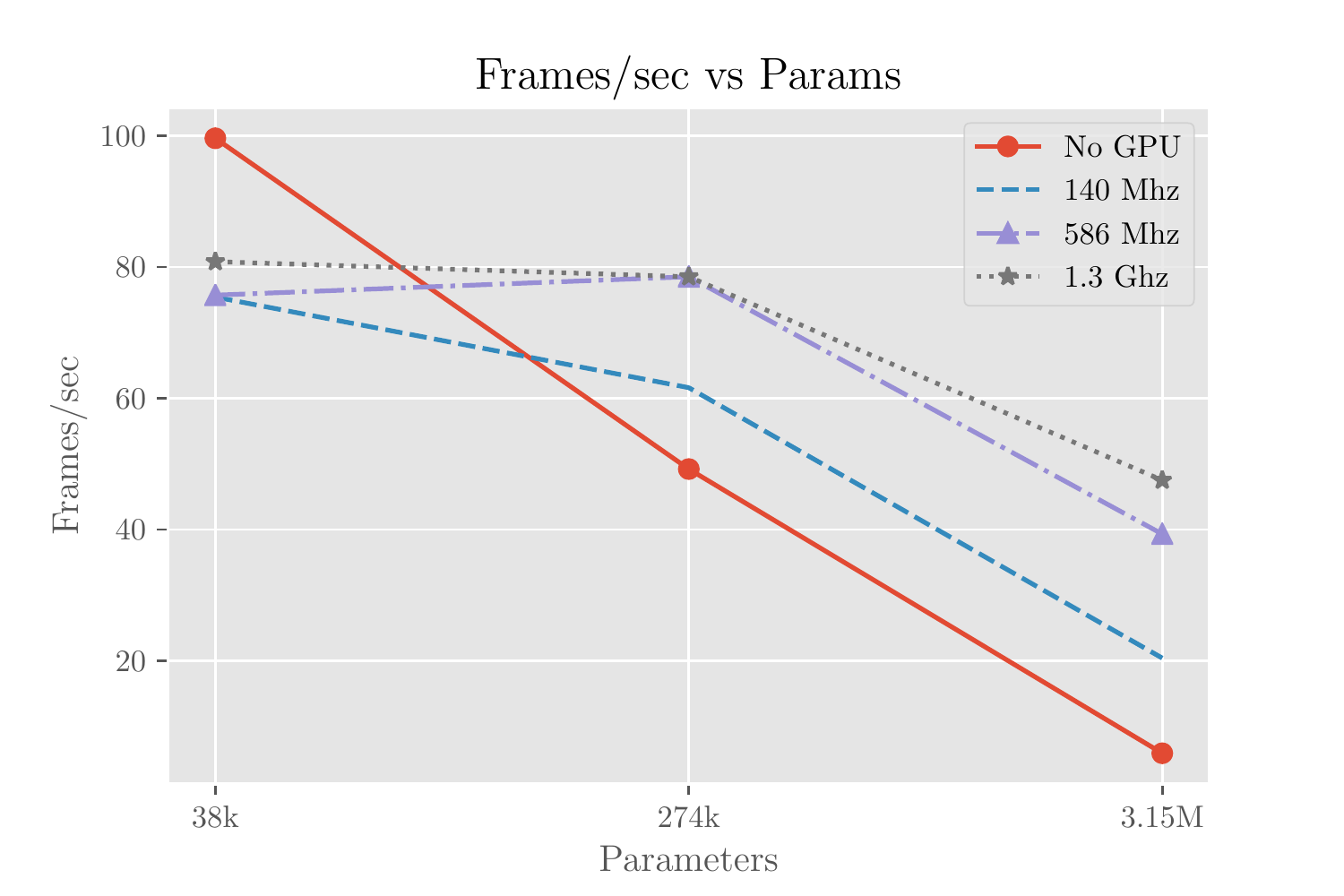}}\quad
  \subfigure[Performance]{\includegraphics[scale=0.33]{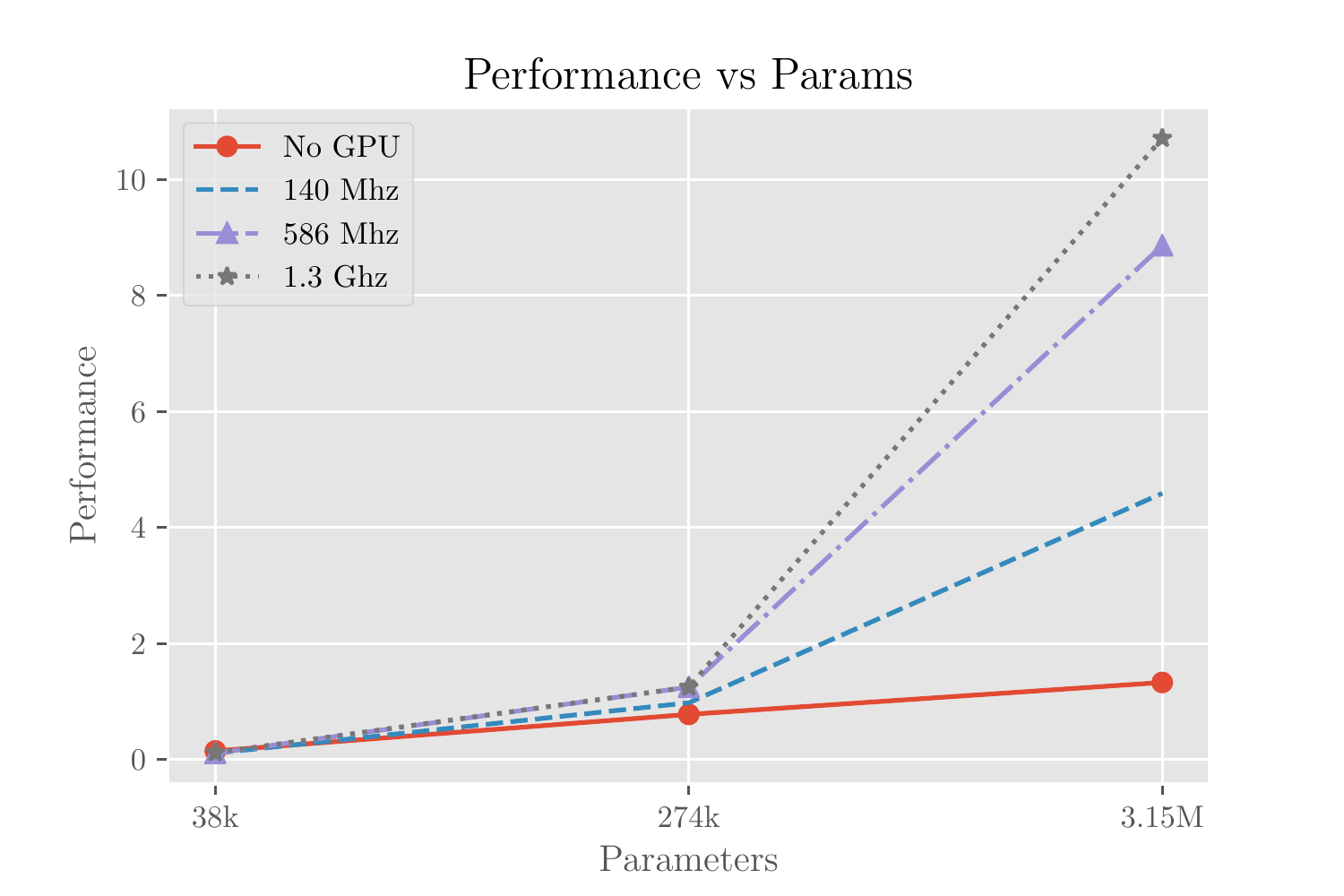}}\quad
  \subfigure[Energy Efficiency]{\includegraphics[scale=0.33]{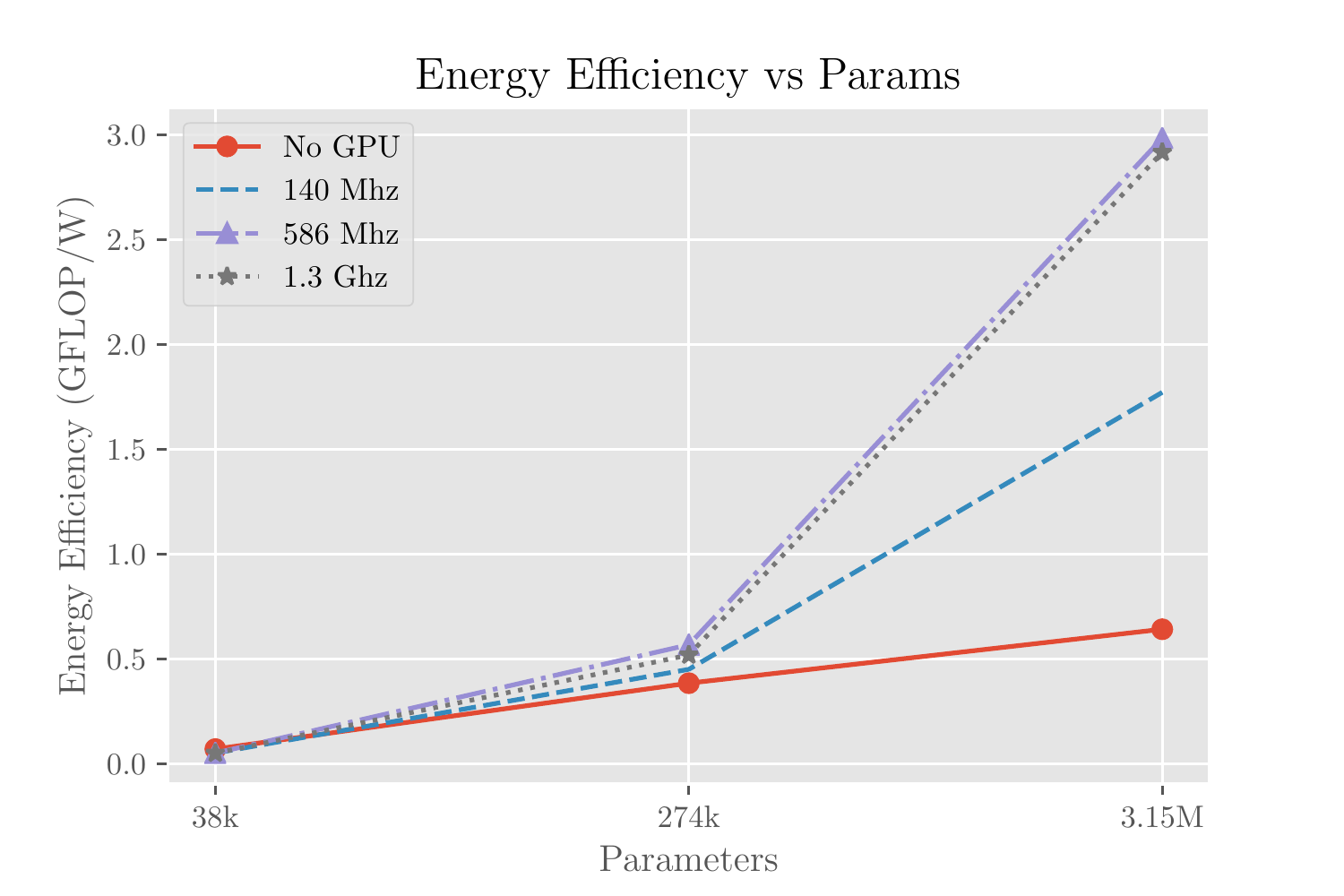}}\quad
    \label{fig:rew}

  \subfigure[Power - Config 1]{\includegraphics[scale=0.33]{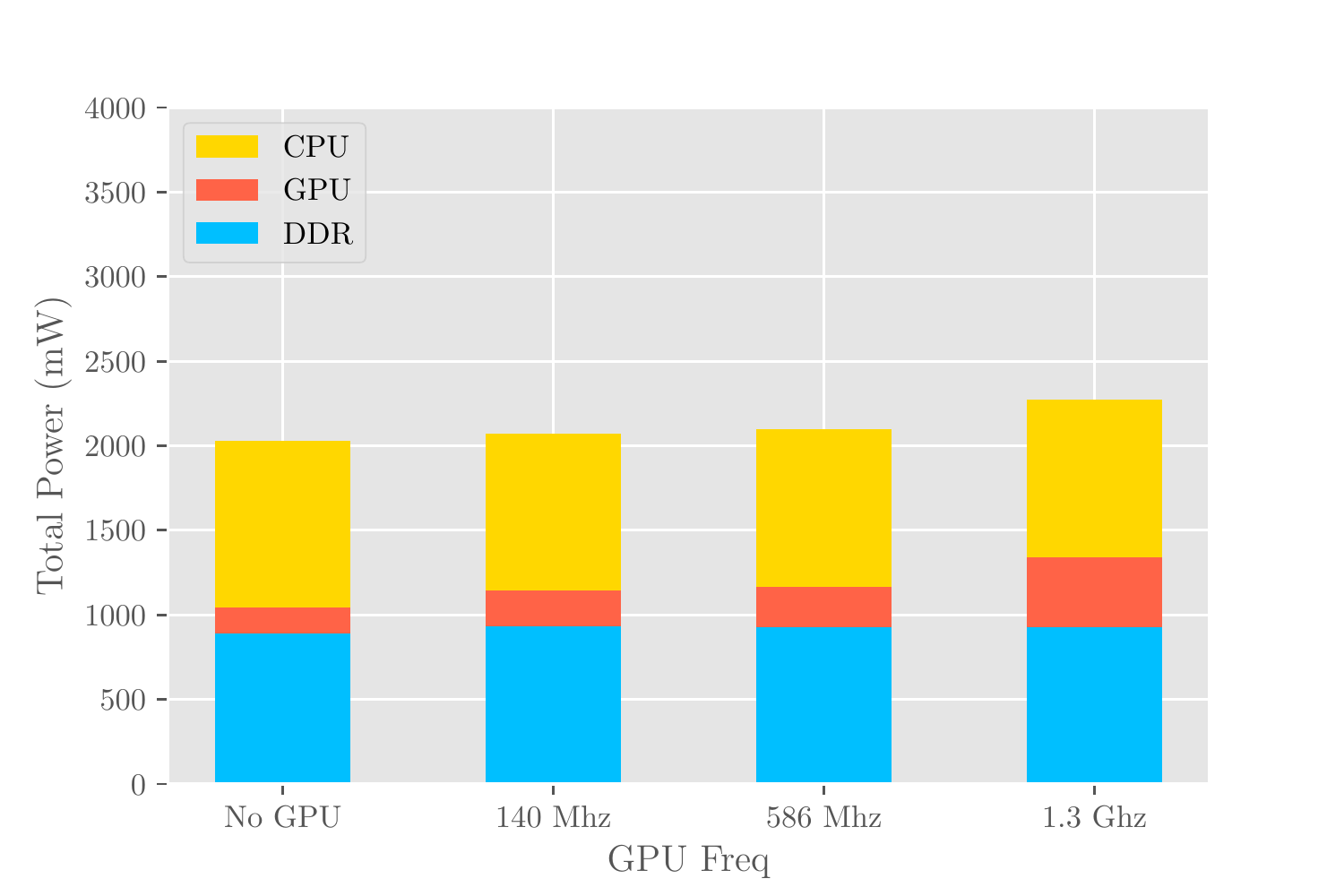}}\quad
  \subfigure[Power - Config 2]{\includegraphics[scale=0.33]{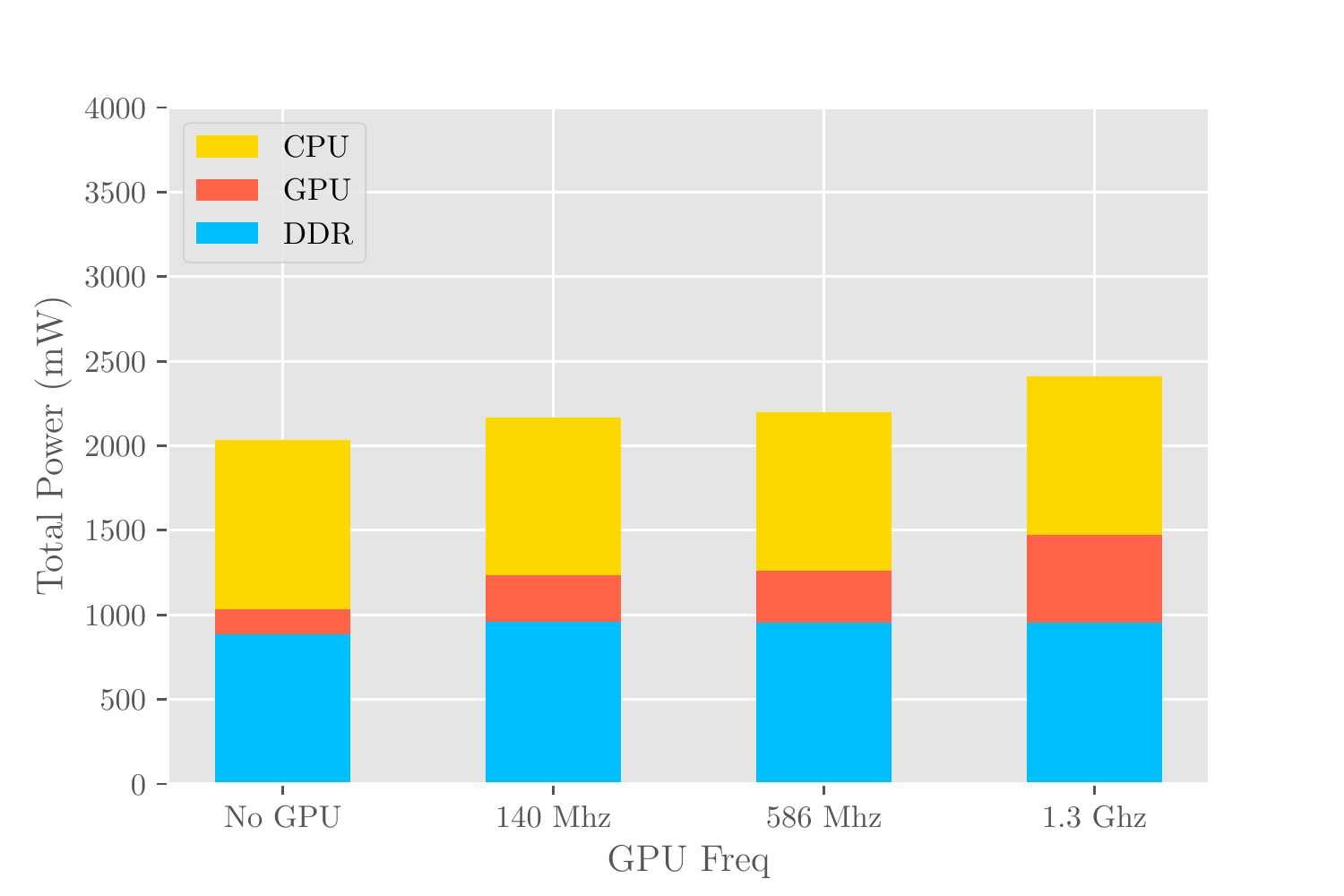}}\quad
  \subfigure[Power - Config 3]{\includegraphics[scale=0.33]{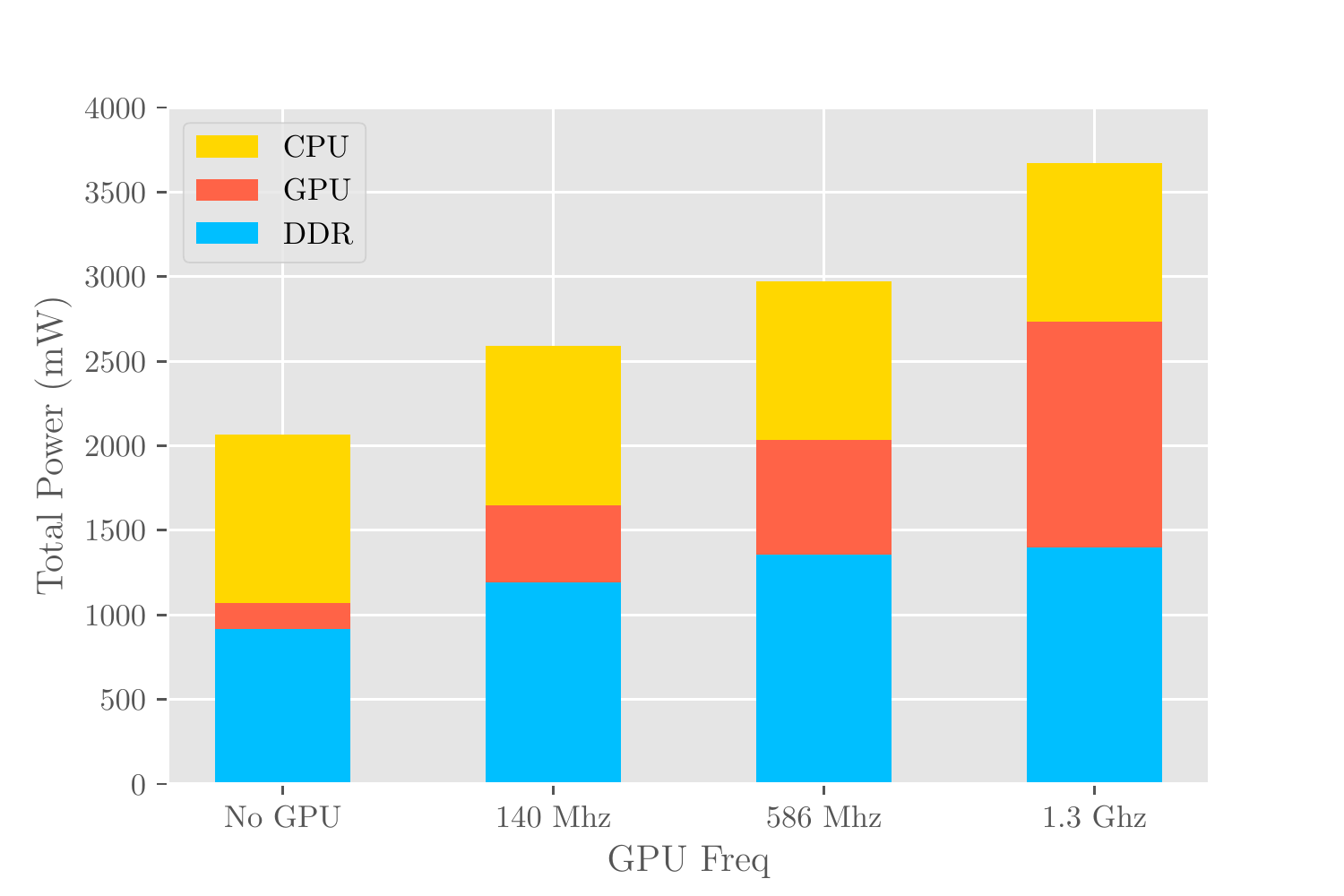}}\quad
    \label{fig:rew}
    
\caption{This figure shows how the Power, Energy, Frame Rate and Performance change for different configurations of the autoencoder}
\label{fig:tx2}
\end{figure*}

We measured power consumption and energy efficiency of these models on the Embedded Nvidia TX2 platform in order to demonstrate the feasibility of such architectures on embedded devices. We performed tests on the TX2 with different combinations of GPU and CPU frequencies. Figure \ref{fig:tx2} shows these metrics where we measure the power, energy efficiency, frame rate and performance for the three autoencoder configurations.  As shown in Figure \ref{fig:tx2}, even the largest configuration that we tested consumes about 3 watts at the 586 MHz GPU frequency, while being able to process more than 60 frames per second.  This frame rate is more than sufficient for most embedded navigation and control tasks, and 3 watts of power is a small percentage of the power consumed by electric cars and drones, which are often targeted platforms for autonomous embedded systems.  

Additionally, we conducted similar hardware experiments
with config 2 to find the CPU/GPU frequencies that result in optimal implementation.
We choose Config 2 because the RL agent performance of Config 2 and Config 3 are very similar even though Config 2 has significantly less parameters. Table \ref{fig:config2hw} shows the Power, frame rate and energy efficiency for the various combinations of CPU and GPU frequencies on the NVIDIA TX2. In the experiments that we did, the frame rate requirements were fixed to be 60 FPS. From Table \ref{fig:config2hw} we see that a few combinations are able to achieve this frame rate. It also shows 2 combinations where no GPU were used. It either failed to achieve 60 FPS or the power consumption and energy efficiency was above 3 W. The ideal combination we found was CPU 1.4 GHz and GPU 1.3 GHz/586 Mhz which 
are among those that meet the frame rate of 60 FPS and consume 1.8 W or less. 


\begin{table*}[]
\resizebox{.92\linewidth}{!}{
\begin{tabular}{c|c|c|c|c|c|c|c|c|c|c}
\toprule
\multirow{2}{*}{CPU Freq} & \multirow{2}{*}{GPU Freq} & \multicolumn{4}{c|}{Power (mW)}      & \multirow{2}{*}{Time/frame(ms)} & \multirow{2}{*}{Frame Rate} & \multirow{2}{*}{\begin{tabular}[c]{@{}l@{}}Performance\\ (GFLOP/S)\end{tabular}} & \multirow{2}{*}{Energy(J)} & \multirow{2}{*}{\begin{tabular}[c]{@{}l@{}}Energy Eff\\ (GFLOPS/W)\end{tabular}} \\
                          &                           & CPU   & GPU   & DDR   & Total  &                             &                      &                          &                         &                             \\
\midrule
\multirow{3}{*}{345. MHz} & 140 MHz                   & 230   & 198.5 & 372.5 & 801.0  & 72.9                        & 13.7                 & 0.219                    & 0.058                   & 0.273                       \\
                          & 586 MHz                   & 230   & 199.4 & 804.7 & 1234.1 & 65.1                        & 15.4                 & 0.245                    & 0.080                   & 1.98                        \\
                          & 1.3 GHz                   & 230   & 33.5  & 804.1 & 1369.5 & 62.7                        & 15.9                 & 0.254                    & 0.086                   & 0.186                       \\
\midrule
\multirow{3}{*}{1.4 GHz}  & 140 MHz                   & 537   & 244.2 & 917.6 & 1698.8 & 19.4                        & 51.6                 & 0.823                    & 0.033                   & 0.484                       \\
                          & \textbf{586 MHz}          & \textbf{537}   & \textbf{261.8} & \textbf{915.5} & \textbf{1714.3} & \textbf{16.8}                      & \textbf{59.6}                & \textbf{0.949 }                  & \textbf{0.029 }                  & \textbf{0.554 }                      \\
                          & \textbf{1.3 GHz}          & \textbf{537}   & \textbf{416.1} & \textbf{917.8} & \textbf{1870.9} & \textbf{16.7}                        & \textbf{60.1}                 & \textbf{0.957}                    & \textbf{0.031}                  & \textbf{0.512}                       \\
\midrule
\multirow{3}{*}{2 GHz}    & 140 MHz                   & 931.8 & 288.5 & 966.2 & 2186.5 & 16.9                        & 59.1                 & 0.941                    & 0.037                   & 0.431                       \\
                          & 586 MHz                   & 935.8 & 323.3 & 955.9 & 2214.9 & 12.9                        & 77.4                 & 1.234                    & 0.029                   & 0.557                       \\
                          & 1.3 GHz                   & 933.1 & 545.6 & 953.9 & 2432.6 & 12.7                        & 78.7                 & 1.255                    & 0.031                   & 0.516                       \\
\midrule
2 GHz                     & \multirow{2}{*}{No GPU}   & 997.0  & 153   & 882.5  & 2032.5 & 20.3                        & 49.2                 & 0.785                    & 0.041                   & 0.386                       \\
2 GHz (4 cores)           &                           & 2158.3 & 178.5 & 1031.1 & 3367.9 & 15.8                        & 63.4                 & 1.016                    & 0.053                   & 0.302                    \\
\bottomrule
\end{tabular}
}
\caption{Hardware performance of Config 2 with different combinations of CPU and GPU frequencies}
    \label{fig:config2hw}
\end{table*}

\section{Conclusions and Future Works}

Through our experiments, we have demonstrated the usefulness of autoencoders for compression in embedded reinforcement learning tasks.  Using this architecture, we can significantly reduce the time needed to produce a viable policy while also preserving a sufficiently high throughput and low power to perform a large number of embedded reinforcement learning tasks.  There are significant steps that one could take to further improve the sample efficiency and safety of these methods.  For example, one could use the compressed representation generated by the variational autoencoder to learn to model the dynamics on an environment, and subsequently train a policy completely in a simulation, further increasing the safety of the training process \cite{ha2018world}.  Additionally, fear based methods provide a way to reduce dangerous environment interactions by using human inputs \cite{Saunders2017}.  We chose to use a more simple reinforcement learner because it would be the most broadly applicable method and thus a good way to compare different autoencoder configurations.  However, the incorporation of these methods could significantly increase sample efficiency in some cases. Future work will also include the deploying and testing of our embedded models on robotic systems as quadrotors or ground robots.




\section{ Acknowledgments}
This project was sponsored by the U.S. Army Research Laboratory under Cooperative Agreement Number W911NF-10-2-0022. The views and conclusions contained in this document are those of the authors and should not be interpreted as representing the official policies, either expressed or implied, of the U.S. Government. The U.S. Government is authorized to reproduce and distribute reprints for Government purposes notwithstanding any copyright notation herein.


%

%
\appendix

\end{document}